\documentclass[12pt]{amsart}

\usepackage[russian,english]{babel}
\usepackage{amscd}
\usepackage{amsmath}
\usepackage{amsfonts}
\usepackage{amssymb}
\usepackage{graphics}
\usepackage{epsfig}
\usepackage{array} 

\usepackage{listings}
\usepackage{xcolor}
\lstdefinestyle{sharpc}{language=[Sharp]C, frame=lr,
rulecolor=\color{blue!80!black}}

 2

\newcommand{\CS}{C\nolinebreak\hspace{-.05em}\raisebox{.6ex}{\scriptsize\bf \#}}

\numberwithin{equation}{section} \numberwithin{theorem}{section}

\title[The Algorithmic Inflection of Russian]{The Algorithmic Inflection of Russian and Generation of Grammatically Correct Text}

\author{T.M.\,Sadykov and T.A.\,Zhukov}
\address{Department of Mathematics
\newline \indent and Computer Science,
\newline \indent Plekhanov Russian University
\newline \indent 115054, Moscow, Russia}
\email{Sadykov.TM@rea.ru}
\thanks{This research was conducted in the framework of the basic
part of the scientific research state task in the field of
scientific activity of the Ministry of science and education of
the Russian Federation, project no.~2.9577.2017. The first author
was also supported by the grant of the Government of the Russian
Federation for investigations under the guidance of the leading
scientists of the Siberian Federal University (contract
No.~14.Y26.31.0006).}

\begin{document}

\begin{abstract}
We present a deterministic algorithm for Russian inflection. This
algorithm is implemented in a publicly available web-service
www.passare.ru which provides functions for inflection of single
words, word matching and synthesis of grammatically correct
Russian text. The inflectional functions have been tested against
the annotated corpus of Russian language
\texttt{OpenCorpora}~\cite{openCorpora}.
\end{abstract}

\maketitle

\section{Introduction}
\label{sec:introduction}

Automatic inflection of words in a natural language is necessary
for a variety of theoretical and applied purposes like parsing,
topic-to-question generation~\cite{Chali-Hasan}, speech
recognition and synthesis, machine
translation~\cite{IomdinOliverSagalova}, tagset
design~\cite{Kuzmenko}, information retrieval~\cite{Iomdin},
content analysis etc~\cite{Belonogov,BelonogovKotov}. Various
approaches towards automated inflection have been used to deal
with particular aspects of
inflection~\cite{EnglishPlural,Zaliznyak} in predefined
languages~\cite{EnglishFrenchSwedish,Foust-AutomaticEnglishInflection,Latin,Sanskrit,RussianAndUkrainian,
Porter} or in an unspecified inflected
language~\cite{languageFree}.

Despite substantial recent progress in the
field~\cite{RussianAndUkrainian,Sorokin,Xiao-Zhu-Liu}, automatic
inflection still represents a problem of formidable computational
complexity for many natural languages in the world. Most
state-of-the-art approaches make use of extensive manually
annotated corpora that currently exist for all major
languages~\cite{Segalovich}. Real-time handling of a dictionary
that contains millions of inflected word forms and tens of
millions of relations between them is not an easy
task~\cite{Goldsmith}. Besides, no dictionary can ever be
complete. For these reasons, algorithmic coverage of the grammar
of a natural language is important provided that inflection in
this language is complex enough.

Russian is a highly inflected language whose grammar is known for
its complexity~\cite{Sorokin,Zaliznyak}. In Russian, inflection of
a word may require changing its prefix, root and ending
simultaneously while the rules of inflection are highly
complex~\cite{Halle-Matushansky,Zaliznyak}. The form of a word can
depend on as many as five grammatical categories such as number,
gender, person, tense, case, voice, animacy etc. By an estimate
based on~\cite{openCorpora}, the average number of different
grammatical forms of a Russian adjective is~11.716. A Russian verb
has, on average,~44.069 different inflected forms, counting
participles of all kinds and the gerunds.

In the present paper we describe a fully algorithmic
dictionary-free approach towards automatic inflection of Russian.
The algorithms described in the present paper are implemented
in~\CS\, programming language. The described functionality is
freely available online at www.passare.ru through both manual
entry of a word to be inflected and by API access of main
functions for dealing with big amounts of data.


\section{Algorithms and implementation}
\label{sec:implementation}

The web-service passare.ru offers a variety of functions for
inflection of single Russian words, word matching, and synthesis
of grammatically correct text. In particular, the inflection of a
Russian noun by number and case, the inflection of a Russian
adjective by number, gender, and case, the inflection of a Russian
adverb by the degrees of comparison are implemented. Russian verb
is the part of speech whose inflection is by far the most
complicated in the language. The presented algorithm provides
inflection of a Russian verb by tense, person, number, and gender.
It also allows one to form the gerunds and the imperative forms of
a verb. Besides, functions for forming and inflecting active
present and past participles as well as passive past participles
are realized. Passive present participle is the only verb form not
currently supported by the website due to the extreme level of its
irregularity and absence for numerous verbs in the language.

The algorithmic coverage of the Russian language provided by the
web-service passare.ru aims to balance grammatical accuracy and
easiness of use. For that reason, a few simplifying assumptions
have been made: the Russian letters "\textcyrillic{\"е}" and
"\textcyrillic{е}" are identified; no information on the stress in
a word is required to produce its inflected forms; for
inflectional functions, the existence of an input word in the
language is determined by the user. Furthermore, the animacy of a
noun is not treated as a variable category in the noun-inflecting
function despite the existence of~1037 nouns (about~1.4\% of the
nouns in the \texttt{OpenCorpora} database~\cite{openCorpora})
with unspecified animacy. This list of nouns has been manually
reviewed on a case-by-case basis and the decision has been made in
favor of the form that is more frequent in the language. The other
form can be obtained by calling the same function with a different
\texttt{case} parameter (\texttt{Nominative} or \texttt{Genitive}
instead of \texttt{Accusative}).

Similarly, the perfectiveness is not implemented as a parameter in
a verb-inflecting function although by~\cite{openCorpora} there
exist~1038 verbs (about 3.2\% of the verbs in the database) in the
language whose perfectiveness is not specified. For such verbs,
the function produces forms that correspond to both perfective and
imperfective inflections.

The inflectional form of a Russian word defined by a choice of
grammatical categories (such as number, gender, person, tense,
case, voice, animacy etc.) is in general not uniquely defined.
This applies in particular to many feminine nouns, feminine forms
of adjectives and to numerous verbs. For such words, the
algorithms implemented in the web-service passare.ru only aim at
finding one of the inflectional forms, typically, the one which is
the most common in the language.

\begin{figure}[ht!]
\centering{}\includegraphics[width=0.9\textwidth]{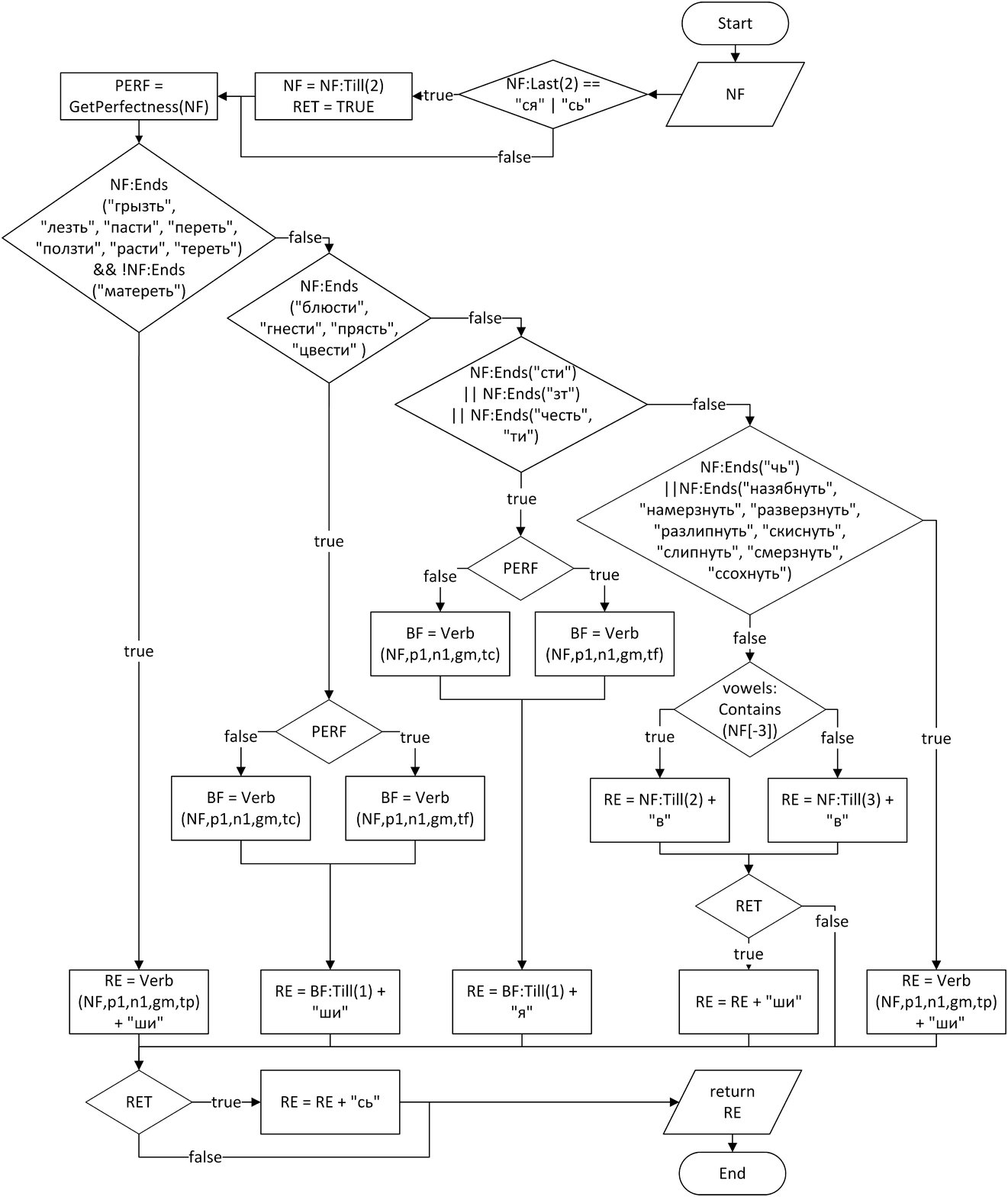}
\caption{Generation of the perfective gerund form of a verb}
\label{fig:perfectiveGerundFlowchart}
\end{figure}

Due to the rich morphology of the Russian language and to the high
complexity of its grammar, a detailed description of the
algorithms of Russian inflection cannot be provided in a journal
paper. The algorithm for the generation of the perfective gerund
form of a verb is presented in
Fig.~\ref{fig:perfectiveGerundFlowchart}. Most of the notation in
Figure~\ref{fig:perfectiveGerundFlowchart} is the same as that of
the~\CS\, programming language. Furthermore, \texttt{NF} denotes
the input normal form (the infinitive) of a verb to be processed.
\texttt{GetPerfectness()} is a boolean function which detects
whether a verb is perfective or not. \texttt{Verb()} is the
function which inflects a given verb with respect to person,
number, gender and tense (see the notation in
Section~\ref{sec:API}). \texttt{BF} denotes the basic form of a
Russian verb which is most suitable for constructing the
perfective gerund of that verb. We found it convenient to use one
of the three different basic forms depending on the type of the
input verb to be inflected. The list \texttt{vowels} comprises all
vowels in the Russian alphabet.

The algorithms have been implemented in~\CS\, programming
language. The implementation comprises about 35,000 lines of code
and has been compiled into a 571~kB executable file.


\section{Software speed tests and verification of results}
\label{sec:speedTests}

The software being presented has been tested against the one of
the widest publicly available corpora of Russian,
\texttt{OpenCorpora}~\cite{openCorpora}. We have been using Intel
Core i5-2320 processor clocked at 3.00GHz with 16GB RAM under
Windows~7. The results are summarized in the below table.

\medskip

\noindent
\begin{center}
\begin{minipage}{17cm}
\centerline{\small Table~1: Inflection speed and agreement rates
of passare.ru and \texttt{OpenCorpora}}
\label{tab:timesOfInflection} {\tiny \vskip0.2cm \noindent
\begin{tabular}{|m{2.1cm}|m{2.9cm}|m{2.1cm}|m{3cm}|m{2.1cm}|m{1.6cm}|}
\hline
{\bf Part of speech } & Total number of words & Total processing time, min:sec & Number of forms computed (per word) & Processing time per word, msec & Agreement rate with \texttt{OpenCorpora}\\
\hline Noun           & 74633                 &        02:36                    & 12    & 2      &  98.557\,\%    \\
\hline Verb           & 32358                 &        05:49                    & 24    & 10     &  98.678\,\%    \\
\hline Adjective      & 42920                 &        00:06                    & 28    & 0.14   &  98.489\,\%    \\
\hline Adverb         & 1507                  &     $<$00:01                    & 2     & 0.021  &  n/a           \\
\hline Ordinal        & 10000 \mbox{(range~0-9999)}  & 00:30                    & 18    & 3      &  n/a           \\
\hline Cardinal       & 10000 \mbox{(range~0-9999)}  & 00:23                    & 24    & 2      &  n/a           \\
\hline Present participle active  & 16946            & 04:55                    & 28    & 17     &  98.961\,\%    \\
\hline Past participle active     & 32358            & 10:19                    & 28    & 19     &  99.152\,\%    \\
\hline Past participle passive    & 32358            & 10:32                    & 28    & 19     &  94.803\,\%    \\
\hline Gerunds                    & 32358            & 00:23                    & 2     & 0.72   &  99.157\,\%    \\
\hline Verb imperative            & 32358            & 00:42                    & 2     & 1      &  95.327\,\%    \\
\hline
\end{tabular}
}
\end{minipage}
\end{center}

\medskip

All of the words whose inflected forms did not show full agreement
with the \texttt{OpenCorpora} database have been manually reviewed
on a case-by-case basis. In the case of nouns, 26.76\% of all
error-producing input words belong to the class of Russian nouns
whose animacy cannot be determined outside the context
(e.g.~"\textcyrillic{\"еж}", "\textcyrillic{жучок}" and the like).
For verbs, 11.26\% of the discrepancies result from the verbs
whose perfectiveness cannot be determined outside the context
without additional information on the stress in the word
(e.g.~"\textcyrillic{насыпать}", "\textcyrillic{пахнуть}" etc.).

Besides, a substantial number of errors in \texttt{OpenCorpora}
have been discovered. The classification of flaws in
\texttt{OpenCorpora} is beyond the scope of the present work and
we only mention that the inflection of the verb
"\textcyrillic{застелить}" as well as the gerund forms of the
verbs "\textcyrillic{выместить}" and "\textcyrillic{напечь}"
appear to be incorrect in this database at the time of writing.


\section{Synthesis of grammatically correct text}
\label{sec:synthesis}

Using the basic functions described above, one can implement
automated synthesis of grammatically correct Russian text on the
basis of any logical, numerical, financial, factual or any other
precise data. The website passare.ru provides examples of such
metafunctions that generate grammatically correct weather forecast
and currency exchange rates report on the basis of real-time data
available online. Besides, it offers a function that converts a
correct arithmetic formula into Russian text.

The following piece of \CS~code is the core of one of the central
functions which generate a grammatically correct report on
exchange rates of currencies.

\lstset{style=sharpc}
\begin{lstlisting}
[SynthFunction("(Change_in_exchange_rate)")]
        public static LogicSet TrendChange(LogicSet input,
                                        LogicSolver solver)
        {
            TimeSeries ts = new TimeSeries();
            var cc1 = input.ElementAt(0).ToString().Substring(1);
            var cc2 = input.ElementAt(1).ToString().Substring(1);
            string rdt = cc1 + cc2;
            var currency1y =
                Financialdata.GetData(rdt,"CURRENCY","1Y","d,c");
            for (int i = 0; i < currency1y.Count; i++)
            {
             int ri = currency1y.Count - i - 1;
             DateTime d = DateTime.Today.AddDays(-i - 1);
             ts.AddData(new TimeSeriesDataPoint(d,
             double.Parse(currency1y[ri][1].Replace('.',','))));
            }
            var result = ts.BuildMontlyTrends();
            if (result.Count == 0) return new LogicSet();
            var last = result.Last();
            var istp = ts.IsTrendPresent(last,
                        DateTime.Today.AddDays(-1));
            DateTime midtrendtime =
            DateTime.FromBinary((last.tstart.Ticks
                                + last.tend.Ticks) / 2).Date;
            int ttlmonths = DateTime.Today.Month -
                                midtrendtime.Month;
            var _id3 = solver.OpenParamGroup();
            var lset = new LogicSet();
            if (ttlmonths > 0)
            {
              var _id2 = solver.OpenParamGroup();
              var a1 = solver.Construct("number_of_months" +
                                             ttlmonths, _id2);
              var dtest =
              solver.Construct("past_time(_months_ago)",_id3);
              solver.Apply(dtest);
              solver.CollapseLongBranches(dtest);
              solver.CloseParamGroup(_id2);
            }
            else        ............
            solver.CloseParamGroup(_id3);
            return lset;
        }
\end{lstlisting}


\section{Automated API access of main functions}
\label{sec:API}

Although all functions of the website www.passare.ru can be
accessed manually by choosing options and typing words to be
inflected or entering the parameters of a text to be automatically
created, API-based automatic access is enabled to speed up work
with big amounts of data. The details of the API access of main
functions are as follows.

\medskip

\texttt{interface: socket}

\texttt{ip: 46.173.208.127}

\texttt{port: 9999}

\texttt{character encoding: UTF8}

\medskip

To access a function, one needs to connect to the server, send a
query string ending with the zero byte, receive the  response
string ending with the zero byte and close the connection.

The API accessible functions of the website provide inflection of
the following parts of speech:

\begin{itemize}

\item Verbs (\texttt{ru\_verb}) with the arguments: verb (the infinitive); person; number; gender; tense;

\item Nouns (\texttt{ru\_noun}) with the arguments: noun (the singular nominative
form); number; case;

\item Adjectives (\texttt{ru\_adjective}) with the arguments: adjective
(the singular masculine nominative form); number; gender; case;
animacy;

\item Adverbs (\texttt{ru\_adverb}) with the arguments: adverb;
comparative/superlative form;

\item Numerals (\texttt{ru\_numeral}):

Cardinals with the arguments: number; card;

Ordinals with the arguments: number; ordi;

Fractions with the arguments: number (e.g. 1/2); frac.

\item {\bf Do we have API accessible functions for participles and gerund form?}

\end{itemize}

The lists of possible values of the parameters in the above
functions are as follows:

Person: \texttt{p1} - 1st person; \texttt{p2} - 2nd person;
\texttt{p3} - 3rd person.

\medskip

Number: \texttt{n1} - Singular; \texttt{nx} - Indefinite plural;
\texttt{n2} - Plural for numerals like 2, 3, 4, 22, 23, 24, etc;
\texttt{n5} - Plural for numerals like 5, 6, 7, 8, etc.

\medskip

Gender: \texttt{gm} - Masculine; \texttt{gf} - Feminine;
\texttt{gn} - Neuter.

\medskip

Tense: \texttt{tc} - Present; \texttt{tp} - Past; \texttt{tf} -
Future.

\medskip

Case:

\texttt{ci} - Imenitelnyj (Nominative)

\texttt{cr} - Roditelnyj  (Genitive)

\texttt{cd} - Datelnyj    (Dative)

\texttt{cv} - Vinitelnyj  (Accusative)

\texttt{ct} - Tvoritelnyj (Instrumental)

\texttt{cp} - Predlozhnyj (Prepositional)

\medskip

Animacy: \texttt{a} - Animate; \texttt{an} - Inanimate.

Adverb form: \texttt{fc} - Comparative; \texttt{fs} - Superlative.

Type of a numeral: \texttt{card} - Cardinal; \texttt{ordi} -
Ordinal; \texttt{frac} - Fractional.

\medskip

Examples of query strings:

\texttt{ru\_adverb;\textcyrillic{быстро};fc}

\texttt{ru\_verb;\textcyrillic{изучить};p3;n1;gm;tc}

\texttt{ru\_adjective;\textcyrillic{русский};nx;gf;ti;na}

\texttt{ru\_noun;\textcyrillic{язык};n1;cp}

\texttt{ru\_numeral;24;card}

\texttt{ru\_numeral;7;ordi}

\texttt{ru\_numeral;11/12;frac}

\medskip

Example of implementation:

\medskip

\texttt{PHP}

\texttt{\$host = 46.173.208.127;}

\texttt{\$port = 9999;}

\texttt{\$waitTimeoutInSeconds=8;}

\texttt{\$fp=\@fsockopen(\$host,\$port,\$errCode,\$errStr,\$waitTimeoutInSeconds);}

\texttt{if (\$fp)}

\texttt{\{ fwrite (\$fp,
"ru\_noun;\textcyrillic{машина};cr;nx"."$\backslash$0");}

\texttt{\$response = fread(\$fp, 10000);}

\texttt{ echo(\$response); \}}

\texttt{//Output: \textcyrillic{машин}}


\section{Discussion}\label{sec:discussion}

There exist several other approaches towards automated Russian
inflection and synthesis of grammatically correct Russian text,
e.g.~\cite{Kanovich-Shalyapina-RUMORS,RussianAndUkrainian}.
Besides, numerous programs attempt automated inflection of a
particular part of speech or synthesis of a document with a rigid
predefined structure~\cite{Chernikov-Karminsky}. Judging by
publicly available information, most of such program make
extensive use of manually annotated corpora which might cause
failure when the word to be inflected is different enough from the
elements in the database.

The solution presented in this paper has been designed to be as
independent of any dictionary data as possible. However, due to
numerous irregularities in the Russian language, several lists of
exceptional linguistic objects (like the list of indeclinable
nouns or the list of verbs with strongly irregular gerund forms,
see Fig.~\ref{fig:perfectiveGerundFlowchart}) have been composed
and used throughout the code. Whenever possible, rational
descriptions of exceptional cases have been adopted to keep the
numbers of elements in such lists to the minimum.



\begin{thebibliography}{99}
{\small

\bibitem{Belonogov}
G.G.\,Belonogov, A.A.\,Horoshilov, and A.A.\,Horoshilov. {\it
Automation of the English-Russian bilingual phraseological
dictionaries based on arrays of bilingual texts (bilingual)},
Automatic Documentation and Mathematical Linguistics, {\bf 44}:3
(2010), 103-110.

\bibitem{BelonogovKotov}
G.G.\,Belonogov and R.\,Kotov. {\it Automated
Information-Retrieval Systems.} Moscow: Mir, 1971.

\bibitem{Chali-Hasan}
Y.\,Chali and S.A.\,Hasan. {\it Towards topic-to-question
generation,} Computational Linguistics, {\bf 41}:1 (2015), 20p.

\bibitem{Chernikov-Karminsky}
B.V.\,Chernikov and A.M.\,Karminsky. {\it Specificities of
lexicological synthesis of text documents,} Procedia Computer
Science, {\bf 31} (2014), 431-439.

\bibitem{EnglishPlural}
D.\,Conway. {\it An algorithmic approach to English
pluralization,} Proceedings of the Second Annual Perl Conference.
San Jose, California, USA. COPE, D., 2001.

\bibitem{EnglishFrenchSwedish}
D.\,Elworthy. {\it Tagset design and inflected languages,}
arXiv:cmp-lg/9504002v2.

\bibitem{languageFree}
M.\,Faruqui, Yu.\,Tsvetkov, G.\,Neubig, and C.\,Dyer. {\it
Morphological inflection generation using character sequence to
sequence learning,} Human Language Technologies, NAACL HLT (2016),
634-643.

\bibitem{Foust-AutomaticEnglishInflection}
W.D.\,Foust. {\it Automatic English inflection,} Proceedings of
the National Symposium on Machine Translation, UCLA (1960),
229-233.

\bibitem{Latin}
H.\,Fuk\'s. {\it Inflection system of a language as a complex
network,} IEEE Toronto International Conference - Science and
Technology for Humanity (2009), 491-496.

\bibitem{Goldsmith}
J.\,Goldsmith. {\it Unsupervised learning of the morphology of a
natural language,} Computational Linguistics {\bf 27}:2 (2001),
153-198.

\bibitem{Halle-Matushansky}
M.\,Halle and O.\,Matushansky. {\it The morphophonology of Russian
adjectival inflection,} Linguistic Inquiry {\bf 37}:3 (2006),
351-404.

\bibitem{Iomdin}
L.L.\,Iomdin. {\it Natural language processing as a source of
linguistic knowledge,} Proceedings of the International Conference
on Machine Learning, Models, Technologies and Applications (2003),
68-74.

\bibitem{IomdinOliverSagalova}
L.L.\,Iomdin, O.\,Streiter, and I.L.\,Sagalova. {\it Learning
lessons from bilingual corpora: Benefits for machine translation,}
International Journal of Corpus Linguistics {\bf 5}:2 (2000),
199-230.

\bibitem{Kanovich-Shalyapina-RUMORS}
M.I.\,Kanovich and Z.M.\,Shalyapina. {\it The RUMORS system of
Russian synthesis,} Proceedings of the 15th conference on
Computational linguistics - Vol. 1 (1994), 177-179.

\bibitem{Sanskrit}
Kasmir Raja~S.\,V., V.\,Rajitha, and Meenakshi Lakshmanan. {\it
Computational model to generate case-inflected forms of masculine
nouns for word search in Sanskrit E-text,} Journal of Computer
Science {\bf 10}:11 (2014), 2260-2268.

\bibitem{RussianAndUkrainian}
M.\,Korobov. {\it Morphological analyzer and generator for Russian
and Ukrainian languages,} Communications in Computer and
Information Science {\bf 542} (2015), 330-342.

\bibitem{Kuzmenko}
E.A.\,Kuzmenko. {\it Morphological analysis for Russian:
Integration and comparison of taggers}, Communications in Computer
and Information Science {\bf 661} (2017), 162-171.

\bibitem{openCorpora} \texttt{OpenCorpora:} an open corpus of Russian
language, http://www.opencorpora.org/

\bibitem{Porter}
M.F.\,Porter. {\it An algorithm for suffix stripping,} Program
{\bf 14}:3, (1980), 130-137.

\bibitem{Segalovich}
I.\,Segalovich. {\it A fast morphological algorithm with unknown
word guessing induced by a dictionary for a web search engine,}
Proceedings of the International Conference on Machine Learning;
Models, Technologies and Applications (2003), 273-280.

\bibitem{Sorokin}
A.\,Sorokin. {\it Using longest common subsequence and character
models to predict word forms,} Proceedings of the 14th Annual
SIGMORPHON Workshop on Computational Research in Phonetics,
Phonology, and Morphology (2016), 54-61.

\bibitem{Xiao-Zhu-Liu}
T.\,Xiao, J.\,Zhu, and T.\,Liu. {\it Bagging and boosting
statistical machine translation systems,} Artificial Intelligence
{\bf 195} (2013), 496-527.

\bibitem{Zaliznyak}
A.A.\,Zaliznyak. {\it Russian Nominal Inflection.} (Russian)
Nauka, 1967.

} 

\end{thebibliography}
\end{document}